\documentclass[journal,compsoc]{IEEEtran}
\makeatletter
\def\@seccntformatinl#1{\csname the#1dis\endcsname\hskip 1em\relax}
\makeatother

\IEEEoverridecommandlockouts
\usepackage{cite}
\usepackage{amsmath,amssymb,amsfonts}
\usepackage{graphicx}
\usepackage{textcomp}
\usepackage{xcolor}
\def\BibTeX{{\rm B\kern-.05em{\sc i\kern-.025em b}\kern-.08em
    T\kern-.1667em\lower.7ex\hbox{E}\kern-.125emX}}
\usepackage{algorithm,algpseudocode} 
\usepackage{cite}
\usepackage[utf8x]{inputenc} 
\usepackage{amsmath}
\usepackage{epsfig}
\usepackage{grffile}
\usepackage{balance,color}
\usepackage{mathtools, nccmath}

\providecommand{\keywords}[1]{{\textit{Index Terms}}}

\usepackage{bbm}
\DeclareMathOperator*{\minimize}{Minimize}

\newcommand\w{\boldsymbol{w}}
\newcommand \W {\boldsymbol{W}}

\begin{document}

\title{Communication-Efficient Hierarchical Federated Learning for  IoT Heterogeneous Systems with Imbalanced Data }   
\author{ Alaa Awad Abdellatif$^{*}$, Naram Mhaisen$^{*}$, Amr Mohamed$^{*}$,  Aiman Erbad$^{\dag}$, Mohsen Guizani$^{*}$, Zaher Dawy$^{+}$, and Wassim Nasreddine$^{\ddag}$   \\
 $^*$College of Engineering, Qatar University \\
 $^\dag$College of Science and Engineering, Hamad Bin Khalifa University  \\
 $^{+}$Electrical and Computer Engineering Department, American University of Beirut \\
 $^{\ddag}$Department of Neurology, American University of Beirut \\ 
 E-mail: \{alaa.abdellatif, aerbad\}@ieee.org, \{naram, amrm, mguizani\}@qu.edu.qa,   \{zd03, wn13\}@aub.edu.lb   
 } 
\maketitle

\begin{abstract} 

Federated learning (FL) is a distributed learning methodology that allows multiple nodes to cooperatively train a deep learning model, without the need to share their local data. It is a promising solution for telemonitoring systems that demand intensive data collection from different locations while maintaining a strict privacy constraint. 
Due to privacy concerns and critical communication bottlenecks, it can become impractical to send the FL updated models to a centralized server.  
Thus, this paper studies the potential of hierarchical FL in IoT  heterogeneous systems and propose an optimized solution for user assignment and resource allocation on multiple edge nodes. In particular, this work focuses on a generic class of machine learning models that are trained using gradient-descent-based schemes while considering the practical constraints of  non-uniformly distributed data across different users.   
We evaluate the proposed system using two real-world datasets, and we show that it outperforms state-of-the-art FL solutions. In particular, our numerical results highlight the effectiveness of our approach and its ability to provide $4$-$6\%$ increase in the classification accuracy, with respect to hierarchical FL schemes that consider distance-based user assignment.  Furthermore, the proposed approach could significantly accelerate FL training and reduce communication overhead by providing $75$-$85\%$ reduction in the  communication rounds between edge nodes and the centralized server, for the same model accuracy.   

\end{abstract}

\section{Introduction\label{sec:Introduction} }     

The rapid evolution of Artificial Intelligence (AI), Internet of Things (IoT), and  big data is paving the path towards a plethora of interactive applications that can inspire substantial transformations in the industrial services.   
However, these new trends come with several challenges due to the latency and reliability constraints, and the need to process and transmit enormous amount of data, generated by IoT devices, with guaranteed quality of service. 
%
Emerging AI technologies, such as deep learning, can turn the vision of building real-time interactive systems into reality \cite{DNN2018, DNNBook}. For instance, in light of the recent pandemic, deep learning has proven its crucial role in analyzing and understanding the outbreaks evolution. Indeed, efficient deep learning techniques can provide intelligent healthcare services, such as event detection and characterization, real-time remote monitoring, and identification of patients with high mortality risks. However, deep learning relies on the availability of large datasets, provided by medical and IoT devices, for training and improving deep learning models \cite{DeepLearning2016}. This puts a major challenge on maintaining data privacy, since medical devices are typically collecting private data. Hence, it is commonly not a practical solution to forward this data to a centralized entity for conducting the training \cite{Felix2020}. Centralized processing of massive amount of data puts users' privacy at high risk, while imposing enormous load on wireless networks.    
Thus, we envision that bringing the intelligence close to the IoT devices, using  Multi-access Edge Computing (MEC) \cite{SurveyMEC2017} along with Federated Learning (FL) \cite{Wang2019}, is a key for supporting data privacy, while jointly training large deep learning models.   
 
FL is a promising solution for training deep learning models without storing the data at a centralized location. 
It allows multiple entities/users to jointly train a deep learning model on their combined data, without revealing their data to a centralized entity. 
This approach of privacy-preserving cooperative learning is carried out by three main steps: 
(i) all participating users receive the latest global model $\W$ from the centralized server (also called a broker); (ii) they train the received model using their local data; and (iii) they upload their locally trained models $\W_i$ back to the centralized server to be aggregated and form an updated global model.  
These steps are repeated until a certain convergence criterion is obtained.  

Following FL protocol, the local devices (or users) never transfer their local data, since only the updated models are communicated.  
However, a major challenge in centralized FL is the massive communication overhead that arises from sending the End Users (EUs) their full models' update to the centralized server at each iteration. Such updates are of the same size as the trained model, which can be in the range of gigabytes for deep learning models with thousands of parameters  \cite{DenselyConvNetwork}.  
Moreover, frequent exchange of the updated models with the centralized server leads to increasing the FL convergence time, especially with non-Independent and Identically Distributed (non-IID) data \cite{Convergence_FedAvg}. 
Accordingly, in large scale networks, centralized FL can generate high communication overheads (in terms of latency, network bandwidth, and energy consumption), which turns the FL to be unproductive in resource-constrained environments.  

\begin{figure}[t!]
	\centering
		\scalebox{1.80}{\includegraphics[width=0.27 \textwidth]{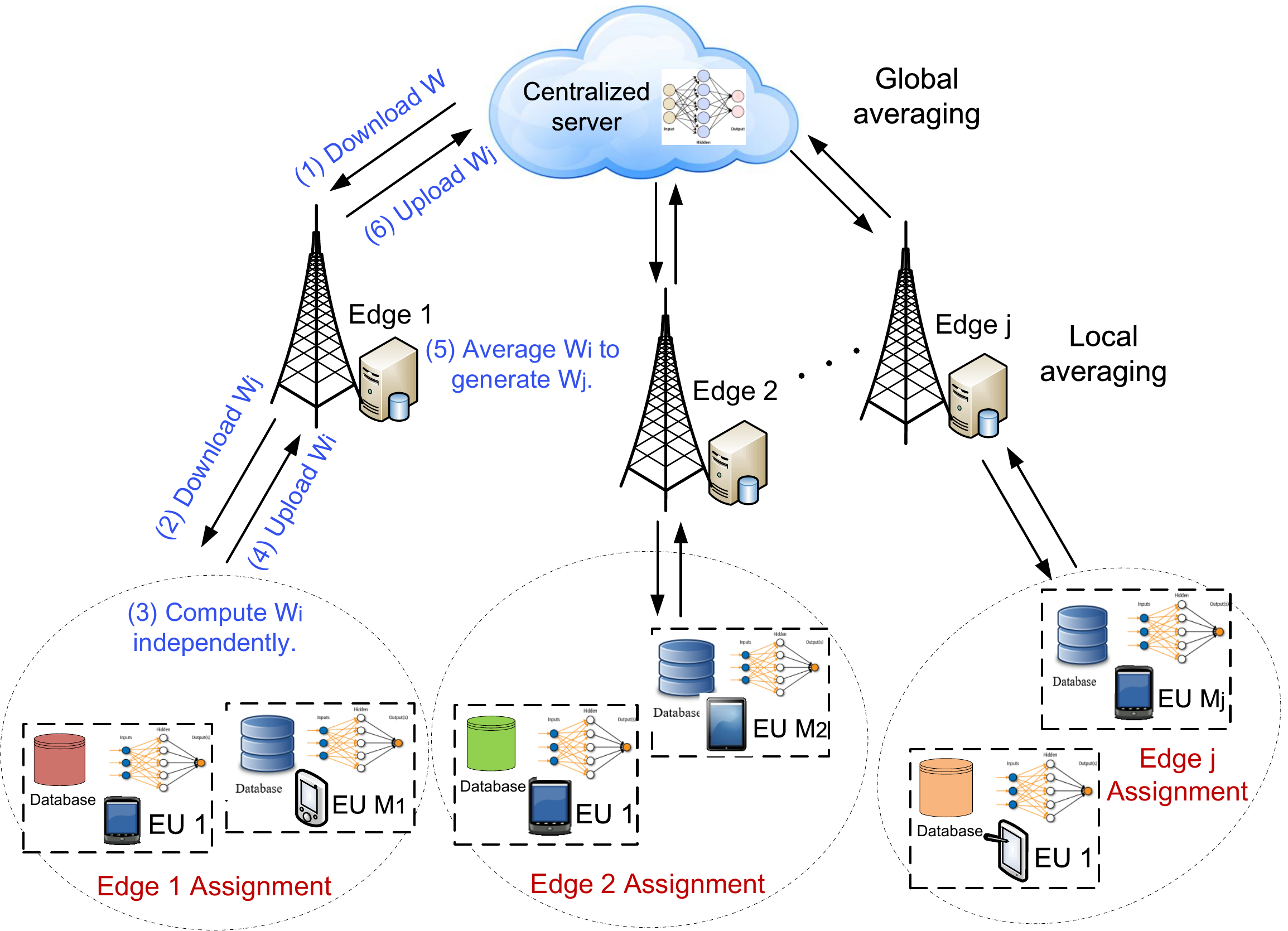}}
	\caption{Hierarchical federated learning: architecture and data flow. }
	\label{fig:HFL}
\end{figure}

To address the above challenges, we propose an optimized EUs assignment and resource allocation scheme over hierarchical federated learning architecture (see Figure~\ref{fig:HFL}), in which multiple edge servers are employed to reduce the overhead resulting from information exchange between EUs and the centralized server. In the proposed hierarchical architecture, EUs are assigned to different edge nodes not only based on their locations (like the state-of-the-art solutions), but also based on their data distributions and communication constraints.  
Based on this, the EUs send their local updates to the selected edge node (e.g., small cell base station) for aggregation. Then, the edge node computes the aggregate model and transfers it back to their assigned EUs to update their models accordingly. After a specific number of  iterations, the edge nodes forward their models' updates to the centralized server to maintain a global model.  
In this context, we propose a novel EUs assignment and resource allocation scheme that aims to minimize the FL convergence time, hence the consumed resources, by optimally assigning the EUs with different data distributions to the available edge nodes.  

Without loss of generality, this paper considers, as a case study, applying the proposed hierarchical FL in large-scale intelligent healthcare systems.     
These systems include thousands of participants who need continuous monitoring at the same time, especially in case of pandemics.  This entails major challenges and puts significant load on different healthcare entities.  
Thus, the strategic solution is to move large number of patients with mild symptoms into remote monitoring or home care.   
Deep learning models can enable home care services by acquiring and processing the patients' data to identify the patients' states. 
However, collecting and forwarding such private data from the patients to a central entity come at a risk of violating the patients' privacy.   
Thus, we envision that leveraging  the proposed hierarchical FL within large-scale intelligent healthcare systems  is a key for enabling remote monitoring.  
In this context, the main contributions of our work can be summarized as:
\begin{enumerate}
	\item Design a distributed learning system with hierarchical FL for supporting intelligent healthcare systems. In the hierarchical FL setting, the EUs are collaborating to train a deep learning model, without exchanging raw data, by connecting to the local edge nodes in their communication range, which are then connected to a centralized server. 
	\item Study the global imbalanced data effect on the obtained accuracy and convergence time of deep learning models trained by hierarchical FL. This includes formulating an optimization problem for EUs assignment and resource allocation based on EUs' data distributions, while considering  latency, bandwidth, and energy consumption constraints.  
	\item Solving the formulated problem using an efficient algorithm that considers both data distributions of different EUs and wireless environment conditions.   
	\item Evaluating the performance of the proposed approach via comprehensive experiments using real-world datasets,  which confirms that our approach provides near-optimal performance for different data distributions and system configurations. 
\end{enumerate}


In the rest of the paper, Section \ref{sec:Related} presents the related work on single-layer FL and hierarchical FL. Section \ref{sec:Sec2} introduces the main challenges that will be tackled in this paper, and presents the considered intelligent  healthcare system architecture with hierarchical FL.  Section \ref{sec:sec3}  introduces first the mathematical modeling of hierarchical FL. Then, it presents the communication and computation latency as well as energy consumption analysis of the proposed hierarchical FL framework.  
Section \ref{sec:sec4} presents the formulated optimization framework along with the solution for EUs assignment and resource allocation over hierarchical FL.  
Performance evaluation of our system is then discussed in Section \ref{sec:sec5}.  Finally, the paper is concluded in Section \ref{sec:conclusion}. 

\section{Related Work\label{sec:Related}}

The proposed FL schemes in the literature can be broadly classified, based on the system architecture, into two categories: Single-layer FL  and hierarchical FL.     

\subsection{Single-layer Federated Learning \label{sec:Single}}

The concept of FL was first proposed in \cite{synchFL}, with its efficiency demonstrated through experiments on different datasets.  
The presented model in \cite{synchFL} considered a single-layer FL, where the users exchanged their updated models with a centralized server that aggregated and formed an updated global model with a fixed frequency.  
This was followed by several extensions, which can be categorized into: 
\begin{itemize}
	\item analyzing the convergence of distributed gradient descent and federated learning algorithms from a theoretical perspective, and optimizing the learning process given computation and communication resource budgets  \cite{zhao_federated_2018, Convergence_FedAvg, Wang2019, WSaad1};  
	\item selecting the users participating in the synchronization process of FL approaches \cite{Nishio2019ClientSF, wang_optimizing_2020, Vincent2020, 9145182};  
	\item developing communication-efficient techniques to reduce the amount of exchanged data in FL systems by adopting various sparsification and compression techniques  \cite{Sparse_Communication, Felix2020, FL_IOT}.  
\end{itemize} 

The effect of non-IID on the performance of FL has been studied in \cite{zhao_federated_2018}. It has been shown, theoretically and empirically, that highly skewed non-IID data can significantly reduce the accuracy of the global learning model by up to $55\%$. 
As a solution to enhance the training on non-IID data, the authors proposed to share globally a small subset of data between all users. Combining this data with the local data of each user turns it to be less biased or skewed. However, exchanging data between different users is not always feasible due to the privacy constraint and communication overhead.  
In \cite{Convergence_FedAvg}, the authors analyzed the convergence rate of Federated Averaging (FedAvg) on non-IID data for strongly convex and smooth problems. 
In \cite{Wang2019}, the authors studied the adaptation of global aggregation
frequency for FL, while considering a fixed resource budget.   
They analyzed the convergence bound of gradient-descent based FL on non-IID data from a theoretical perspective. Then, they used this convergence bound to build a control algorithm that adapts the frequency of global aggregation in real time to minimize the learning loss under a fixed resource budget. 
In \cite{WSaad1}, the convergence analysis of FL within an Unmanned Aerial Vehicle (UAV) swarm was studied. Then, a joint power allocation and scheduling problem was formulated to optimize the convergence rate of FL while considering the energy consumption and delay requirements imposed by the swarm's control system.    

It is shown in \cite{Convergence_FedAvg} that the participation of all users in the FL process forces the central server to wait for \textit{stragglers},  i.e., users who have low-quality wireless links that can significantly slow down the FL process, which turns the FL to be unrealistic.   
Thus, to mitigate the impact of \textit{stragglers}, the authors in  \cite{Nishio2019ClientSF} proposed a method to select a subset of users  for the FL synchronization (or aggregation) process in a resource-constrained environment. They demonstrated the advantages of such technique on improving the FL learning speed.  
This work has been extended in \cite{wang_optimizing_2020}, where a control scheme is proposed, based on reinforcement learning, to accelerate the FL process by actively selecting the best subset of users in each communication round that can counterbalance the bias introduced by non-IID data.  
In \cite{Vincent2020}, a realistic wireless network model was considered to study the convergence time of FL. Then, given the limited wireless resources, a joint learning, resource allocation, and user selection optimization problem was formulated to minimize the FL convergence time and training loss.    
{
In \cite{9145182}, a joint optimization framework for sample selection and user selection was studied to keep a balance between the model accuracy and cost.  However, the distribution distance between different users was optimized in this framework through adjusting the local batch size, which might lead to the under-utilization of data in strongly skewed users. }   

Alternatively, sparsification schemes have been studied to reduce the entropy of the exchanged data (i.e., models' updates) in FL communications.  
The authors in \cite{Sparse_Communication} presented an approach that accelerates the distributed stochastic gradient descent by exchanging sparse updates instead of dense updates. Indeed, they fixed the sparsity rate by only communicating the fraction of entries with the biggest magnitude for each gradient.  
In \cite{Felix2020}, the authors proposed a sparse ternary compression scheme that was designed to maintain the requirements of the FL environment. The proposed scheme  compressed both the upstream and downstream communications of FL leveraging sparsification, ternarization, error accumulation, and optimal Golomb encoding. This study demonstrated the effect of communications compression and data distributions on the obtained performance. However, it neither considered the wireless resources allocation nor the hierarchical FL architecture.   
In \cite{FL_IOT}, the FedAvg scheme was adjusted to use a distributed form of Adam optimization, along with the sparsification and quantization, in order to propose a communication-efficient FedAvg.  

\subsection{Hierarchical Federated Learning \label{sec:Hierarchical}}

Few studies have been proposed so far to address the problem of non-IID data on hierarchical FL architecture. For instance, the authors in \cite{Client-Edge-Cloud} extended the work in \cite{Wang2019} to  analytically prove the convergence of the hierarchical federated averaging algorithm. This work was further extended in \cite{wu_accelerating_2020} considering probabilistic user selection to avoid the impact of \textit{stragglers}.  
In \cite{duan_self-balancing_2021}, a self-balancing FL framework, along with two strategies to prevent the bias of training caused by imbalanced
data distribution, was proposed. The first strategy aimed to perform data augmentation, before training the model, in order to alleviate global imbalance. The second strategy exploited some mediators (which can be considered as edge nodes) to reschedule the training of the users based on the distribution distances between the mediators. 
In \cite{Naram2020}, the effect of the skewed data in hierarchical FL was studied and compared to the centralized FL. Indeed, this work identified the major parameters that affect the learning performance of hierarchical FL. However, this work ignored the resource allocation and wireless communications constraints, such as bandwidth, energy consumption, and latency.     
Although the aforementioned studies have advanced the area of hierarchical FL and user selection from the theoretical perspective, there are still interesting open research directions related to optimized hierarchical FL with wireless resource allocation under practical conditions.   

{\bf Novelty.} Our work is the first to address the problem of EUs' assignment to different edge nodes in hierarchical FL architectures, while considering the users'  data distributions, wireless communication characteristics, and resource  allocation.  
Specifically, unlike other studies, we consider the EUs dataset characteristics and computational capabilities, in addition to practical communication constraints including wireless channel quality, transmission energy consumption, and communication latency.  
Furthermore, the proposed framework for hierarchical FL is incorporated in an intelligent  healthcare system and evaluated using two real-world health datasets to provide efficient remote monitoring  services.

\section{Hierarchical Federated Learning  \label{sec:Sec2}}

In this section, we first highlight the key challenges of using FL in large-scale healthcare systems, then we present our Intelligent-Care (I-Care) system architecture that will be leveraged to address these challenges.    

\subsection{Challenges of FL in healthcare systems \label{sec:why}} 

Before proceeding with the discussion of the proposed framework, we highlight the uniqueness of FL compared to distributed training schemes.  
FL enables a large amount of EUs to jointly learn a model without sharing their local data. Hence, the distribution of both training data and computational resources is fixed and linked to the learning environment. 
However, for efficient leveraging of FL within large scale healthcare systems, the following challenges have to be adequately addressed. 
We remark that these challenges are present in most of the FL scenarios,  however, they are particularly amplified in health-related use-cases. 
   
\textbf{Imbalanced and non-IID data:}   
Given the heterogeneity of healthcare systems, the distribution and size of the acquired datasets at each EU/patient significantly vary. Typically, the training data in FL is collected by diverse IoT devices near/attached to
the EUs, which results in non-IID data distribution over different EUs. 
FedAvg, the leading FL algorithms for non-IID data distributions, suffers from  the large number of communications rounds needed between EUs and the centralized server, especially in the case of imbalanced data \cite{Convergence_FedAvg}.      

\textbf{Large number of participants:} 
Given the evolution of healthcare systems with the increasing number of the chronically ill and elderly people, most of the hospitals are required to serve hundreds of patients daily (especially, in case of pandemic such as the recent COVID-19 outbreak). This puts a significant load on different health entities. A promising solution to face such a demand is to move a large number of patients with mild conditions to home care, while still being monitored remotely.  
This leads to a wide extension in the FL environment for large-scale healthcare systems.    


\textbf{Limited resources:}  
Given the large number of EUs participating in FL, the availability of the needed  computational and communication resources at the EUs and network nodes can be  challenging. The generated traffic for the models' update grows linearly with the number of participating EUs. Hence, distributing the available bandwidth on different EUs in a communication-efficient way is not an easy task. Moreover,  the heterogeneity of the EUs, in terms of computational capabilities and energy availability (i.e., battery level) introduces an extra constraint to optimize the latency and energy consumption. 


\subsection{I-Care system architecture \label{sec:Architecture}} 

\begin{figure}[t!]
	\centering
		\scalebox{1.8}{\includegraphics[width=0.27 \textwidth]{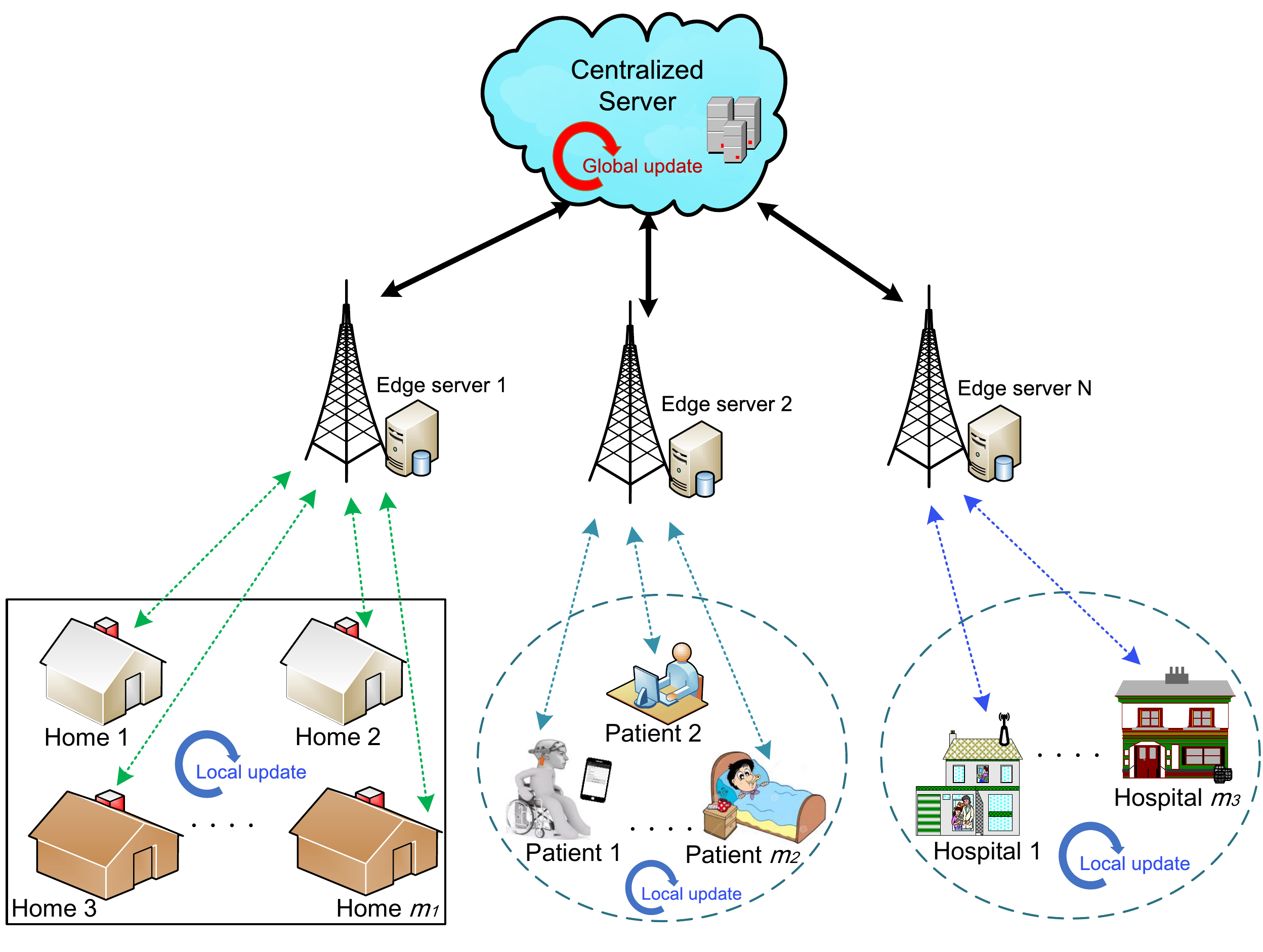}}
	\caption{The proposed I-Care system architecture. }
	\label{fig:system_model}
\end{figure}

For supporting intelligent healthcare services while addressing the above challenges, we consider the I-Care system architecture presented in Figure~\ref{fig:system_model}. The proposed architecture stretches from the EU layer, where data is generated/collected, to the centralized server (or cloud) layer. It  includes the following major components: 

\textbf{EU layer:}   
It is considered that each patient has a combination of IoT devices, which enable real-time monitoring and ubiquitous sensing for his/her health conditions and activities within the smart assisted environment. Examples include: body area sensor networks (i.e., implantable or wearable sensors that measure different bio-signals and vital signs), IP cameras, and external medical and non-medical devices. These IoT devices are connected with a local hub, which is responsible for: 1) gathering the health-related information, 2) processing these data, and 3) training the local learning model at the EU level. 
In our architecture, the EU can be a patient, in-home healthcare environment, or a hospital with several patients. In both cases, FL allows for cooperative training (at the patients or hospital level) without sharing any privacy-sensitive data. 

\textbf{Edge node layer}:  This layer  is an intermediate layer between EUs and the centralized server  that aims to reduce the massive communication overhead that arises from sending full EU model updates to the centralized server.  
In the proposed architecture, the EUs are assigned to different edge nodes, such that each group of EUs runs, with its serving edge node, Synchronous distributed Gradient Descent learning algorithm, also referred to as FedSGD\footnote{FedSGD is a gradient-based algorithm that is equivalent to federated learning when the global aggregation occurs at every step, where every local step is a full deterministic gradient. }  \cite{FedSGD}.  
Under this learning model, an EU $i$ updates its local model $ \W_i$ based on its local dataset $D_i$. Then, the EUs' models are aggregated at the assigned edge node $j$ to form the edge model $\bar{ \W}_j$. Then, all edge nodes synchronize their models at the centralized server  through the FedAvg learning scheme; and the process is repeated until the convergence is attained. 

\textbf{Centralized server layer:}  This layer takes the comparative advantages of powerful computing resources  to aggregate, average,  and  update  the  global learning model. Then, it sends back the global model to all participants to complete the FL process.   

\section{Distributed Resources Optimization \label{sec:sec3}} 
 
In this section, we first introduce the mathematical modeling that defines the problem of hierarchical FL with  imbalanced and non-IID datasets.  Then,  we present the communication and computation latency as well as energy consumption analysis of the proposed hierarchical FL framework.  
{ Table \ref{tab:table1} presents the main mathematical symbols that are used in the following sections.   } 

\subsection{ Hierarchical FL problem formulation \label{sec:Preliminary}} 

We consider the hierarchical FL learning system presented in Figure~\ref{fig:system_model}, which consists of one centralized server, a set $\mathcal{N}$ of $N$ edge nodes, and a set $\mathcal{M}$ of $M$ EUs. Each participating EU $i$ stores a local dataset $\mathcal{D}_i$, with a size $D_i$. 
Hence, the \textit{virtual} dataset at an edge node $j$ is $\mathcal{D}^{(j)} ={\cup_{i}\mathcal{D}^{(i,j)}}, \forall i \in \mathcal{M}_j$, and the \textit{virtual} global dataset at the centralized server is $\mathcal{D}={\cup_{j}\mathcal{D}^{(j)}},  \forall j \in \mathcal{N}$. Note that we use the term \textit{virtual} dataset to clarify the analysis, although these datasets are not stored physically on the edge or the centralized server.  

\begin{table}
    \caption{List of symbols used throughout the paper. }
    \label{tab:table1}
    \centering
    \begin{tabular}{c|c} 
            \hline
         \textbf{Symbol}      & \textbf{Meaning}   \\ \hline 
         $\mathcal{N}$ & A set of $N$ edge nodes. \\
         $\mathcal{M}$ & A set  of $M$ EUs. \\
         $\mathcal{D}_i$ & The local dataset stored at EU $i$. \\
         $\mathcal{C}$ & The set of all possible data classes. \\
         $M_j$ & Number of EUs assigned to edge $j$.  \\
         $\sigma_{j}$ & The proportion of edge $j$ dataset to the global dataset.  \\
         $\left| W_i \right|$ & Updated model length in bits.  \\
         $r_{i,j}^u$ & Upload data rate of EU $i$ to edge $j$.  \\
         $r_{j,i}^d$ & Download data rate of EU $i$ from edge $j$.  \\
         $T_i^c$ & Computation time for updating the local model at EU $i$. \\
         $L_{ij}$ & Transmission delay from EU $i$ to edge $j$.  \\
         $B_{ij}$ & Allocated bandwidth from edge $j$ to EU $i$. \\
         $E_{ij}$ & Energy consumed by EU $i$ to upload the updates to edge $j$.  \\
         $\w_{f}$ & Federated weights.  \\
         $\w_c$ & Virtual central weights.  \\ 
         $D_{KL}$ & Kullback–Leibler divergence (KLD).  \\
         $\lambda_{ij}$ & Assignment indicator of EU $i$ to edge $j$.  \\ \hline
    \end{tabular}
\end{table}

The local dataset $\mathcal{D}_i$, at EU $i$, consists of the collection of data samples $ \left\{x_i, y_i \right\}_{i=1}^{D_i}$, where $x_i$ denotes the data acquired by EU $i$ (i.e., features), and $y_i \in \mathcal{C}$ is the associated label, such that $\mathcal{C}$ is the set of all possible labels or data classes.       
Typically, the main task in any learning model is to find the model parameters $\w$ (i.e., weights) that minimize the loss function $\mathcal{L}(\w)$ using the dataset $D_i$. 
We use the widely adopted cross-entropy loss function:  
\begin{align}
\label{eq:cross_ent_loss}
&\mathcal{L}(\w) &=& \mathbb{E}_{y\sim p(y),\boldsymbol{x}\sim q(\boldsymbol{x}|y)} \left[- \mathbbm{1}_{ y=k} \log d_k(\boldsymbol{x};\w) \right] \nonumber \\
&&=&\sum_{k=1}^C -p(y=k) \mathbb{E}_{\boldsymbol{x}\sim q(\boldsymbol{x}|y=k)} \left[ \log d_k(\boldsymbol{x};\w) \right]
\end{align}
Where $\boldsymbol{x}$ is a feature vector of a data point,  $C$ is the number of classes in the learning problem,  $p(\cdot)$ is the global classes distribution (global distribution), $q(\cdot|\cdot)$ is the likelihood function, and $d_k(\boldsymbol{x},\w)$ is the probability of the $k$-th class for input $\boldsymbol{x}$ under parameters $\w$.
Then, the objective of the overall learning process is to reach a set of parameters $\w$ that minimize the loss function across the \textit{virtual} global dataset: 
\begin{equation}
\label{eq:learn_objective}
\displaystyle{\minimize_{\w} \sum_{k=1}^C p(y=k) \mathbb{E}_{\boldsymbol{x}\sim q(\boldsymbol{x}|y=k)} \left[ \log d_k(\boldsymbol{x};\w) \right]}. 
\end{equation}
The loss function in (\ref{eq:cross_ent_loss}) is minimized using the regular gradient descent update:
\begin{equation}
\label{eq:centr_grad_update}
\w_c^t = \w_c^{t-1}  -\delta \nabla_{\w_c}\mathcal{L}(\w_c). 
\end{equation}
We refer to the parameters in (\ref{eq:centr_grad_update}) as the centralized model.

Following the proposed hierarchical FL, each group of EUs is associated with a specific edge. Then, in the first step of the  hierarchical FL procedure, each EU calculates its learning parameters $\w_{i}$ that are optimized based on its dataset $\mathcal{D}_{i}$ using FedSGD.  
Since the \textit{virtual} global data is distributed across different EUs, each EU $i$ locally calculates the loss function over its own dataset to obtain its learning parameters $\w_{i}$. 
\begin{equation}
\w_{i}^t = \w_{i}^{t-1}  -\delta \nabla_{\w_{i}}\mathcal{L}(\w_{i}). 
\end{equation}
where 
\begin{align}
&\mathcal{L}(\w_i) &=& \mathbb{E}_{y\sim p^{i}(y), \boldsymbol{x}\sim q(\boldsymbol{x}|y)} \left[ \mathbbm{1}_{ y=k} \log d_k(\boldsymbol{x};\w) \right] \nonumber \\
&&=&\sum_{k=1}^C p^{i}(k) \mathbb{E}_{\boldsymbol{x}\sim q(\boldsymbol{x}|y=k)} \left[ \log d_k(\boldsymbol{x};\w) \right].
\label{eq:localmodel}
\end{align}
In (\ref{eq:localmodel}),  $p^{i}(\cdot)$ is the classes distribution of EU $i$.  

In the hierarchical FL settings, the synchronization (i.e., aggregation) of EUs' parameters (weights) is periodically done by taking the weighted average of EUs' parameters with respect to their local datasets' size. The EUs then exploit the averaged parameters until the next aggregation round.  
Hence, the second step in the hierarchical FL procedure is to synchronize the local weights  across all EUs belonging to a specific edge node every $T'$ local gradients steps. Hence, the parameters of an edge node $j$ at  $a$-th \emph{edge aggregation} iteration are: 
\begin{equation}
\w^{a}_{j} =\sum_{i=1}^{M_j} \sigma_{i,j}\w_{i}^{a\times T'},
\label{eq:localw}
\end{equation}
where
\begin{equation}
    \sigma_{i,j} =\frac{|\mathcal{D}^{(i,j)}|}{ |\cup_i\mathcal{D}^{(i,j)}|}.
\end{equation} 
In (\ref{eq:localw}), $M_j$ refers to the number of EUs associated with edge node $j$.    
Similarly, the learning models are synchronized across all edge nodes every ($T'\times T$) steps, where $T$ is the centralized aggregation frequency. Hence, at $b-th$ \emph{centralized aggregation}, the learning parameters are averaged across all edge nodes to obtain the federated weights as:   
\begin{equation}
\w_{\text{f}}^{b} =\sum_{j=1}^N  \sigma_{j}\w_{j}^{b \times T}
\end{equation} 
where
\begin{equation}
    \sigma_{j} =\frac{|\mathcal{D}^{(j)}|}{ |\cup_j\mathcal{D}^{(j)}|}. 
\end{equation} 

Finally, the updated model at the centralized server  (i.e., federated weights) is sent back to the edge nodes as well as EUs, and the process is repeated until maintaining the convergence.  
We remark here that the proposed framework considers the synchronous FL approach, which has shown its efficiency in \cite{Wang2019, synchFL, synchFL2016} compared to asynchronous approaches.  
In synchronous FL, the computation steps are synchronized between all EUs, meaning that all EUs have to finish updating their local models before proceeding with the communication step (to forward their models to the edge nodes). 
Instead, asynchronous FL \cite{synchFL2016} is an alternative to the typically used synchronous FL, in which the edge nodes operate in an asynchronous manner. 
The main advantage of asynchronous FL is the ability to fully exploit the available computational resources, at each EU, by performing more gradient descent iterations at powerful (or faster) EUs \cite{Wang2019}. However, in case of imbalanced and/or non-IID datasets, the aggregated models will be biased to the faster EUs,  which may lead to considerable performance degradation.    
On the contrary, the performance of  synchronous FL may be limited to the performance of the worst EU, i.e., a user with the largest computation and communication delay. 
However, this limitation can be addressed using the proposed solution, which aims to optimize the allocated resources for different EUs, such that it guarantees the best possible performance for all EUs (as will be shown in the next section).   

\subsection{Computation and Communication Latency Analysis \label{sec:Latency}} 
 
In the proposed hierarchical FL framework, we also consider both the communication and computation time for optimizing the performance.  
In the local update step of our hierarchical FL (i.e., from the EUs to the edge nodes), we denote the minimum upload rate of EU $i$ to edge $j$ by $r_{i,j}^u$. Hence, the maximum uplink communication latency of models aggregation at different edge nodes is defined as
\begin{equation}
L^u =  \max_{i} \left( \frac{\left| W_i \right|}{r_{i,j}^u} +	\xi_i^u \right), \forall i \in \mathcal{M},  
\label{eq:delay1}
\end{equation}  
where $\left| W_i \right|$ is the total amount of bits that have to be uploaded and downloaded by every EU during the training. In (\ref{eq:delay1}),  $\frac{\left| W_i \right|}{r_{i,j}^u}$ and $\xi_i$ are, respectively, the transmission time and the access channel delay that EU $i$ expects to experience when transmitting $\left| W_i \right|$ bits to edge node $j$.  In other words, it refers to 
the estimated end-to-end delay when using a given technology \cite{8245779}.  
Similarly, if $r_{j,i}^d$ is the minimum download rate from edge node $j$ to EU $i$,  the maximum downlink latency of receiving the updates at the EUs will be 
\begin{equation}
{L}^d =  \max_{j} \left( \frac{\left| W_i \right|}{r_{j,i}^d} +	\xi_j^d \right), \forall j \in \mathcal{N},  
\label{eq:delay2}
\end{equation} 

For  the computation time per local iteration at EU $i$, it depends on the size of the FL model, number of CPU cycles that are needed to execute one sample of data, which is denoted by $\psi_i$, and size of the local dataset $D_i$ at each EU \cite{flOptimizationModel}.  Given that all samples $\left\{x_i, y_i \right\}_{i=1}^{D_i}$  have the same size (i.e., number of bits), the number of CPU cycles needed for EU $i$ to run one local iteration is ($\psi_i \cdot D_i$). 
The total number of iterations needed to update the local model, at EU $i$, is upper bounded by $\mathcal{O}(\log({1}/{\epsilon}))$, where $\epsilon$ is the local accuracy. This upper bound is applicable for a wide range of iterative algorithms, such as gradient descent, coordinate descent, or stochastic dual coordinate descent \cite{iteration}. Moreover, it is proven in \cite{uperBound} that whatever algorithm is used in the computation phase, FL performance will not be affected as long as the convergence time of these algorithms is upper-bounded by $\mathcal{O}(\log({1}/{\epsilon}))$.   
Hence, the computation time for updating the local model at EU $i$ can be written as a function of    
$ T_i^c  = f(\left| W_i \right|, \upsilon, \epsilon,  \psi_i, D_i, f_i) $,  
where $f_i$ is the CPU-cycle frequency, and $\upsilon$ is a constant that depends on the dataset size and conditions of the local problem \cite{iteration}.    
We remark here that our scheme focuses on the delay over the links between EUs and edge nodes, since the delay over the links between edge nodes and the centralized server can be considered constant for different edge node-EU associations.

\subsection{Transmission Energy Consumption Analysis \label{sec:Energy}} 

In the considered hierarchical architecture, it is assumed that the EU can be an IoT device (e.g., a smartphone) with a battery-limited capacity. Hence, it is important to consider the energy consumption that the EU incurs for transferring its local updates to the edge.  
The transmitted energy consumption at EU $i$ is mainly a function of its wireless channel state, transmission rate, and allocated bandwidth.   The wireless channel between an EU $i$ and an edge node $j$ is characterized by the received signal to noise ratio (SNR), denoted by $\gamma_{ij}$, which is defined as 
\begin{equation}
\gamma_{ij} = \frac{P^r_{ij} \cdot {|h_{ij}|}^2}{N_0 \cdot B_{ij}},
 	\label{gamma}
\end{equation}  
where $P^r_{ij}$ is the received power at the edge node, $|h_{ij}|$ is the fading channel magnitude, $N_0$ is the noise spectral density, and $B_{ij}$ is the allocated bandwidth. Also, we consider a deterministic path loss model where the EU and the edge node are separated by a distance $d_{ij}$, hence the received power $P^r_{ij}$ is attenuated with respect to the transmission power $P^t_{ij}$ following $P^r_{ij} = P^t_{ij} \cdot \omega \cdot d_{ij}^{-\alpha}$, where $\alpha$ is the path loss exponent (such that $2 \leq \alpha \leq 6$), and $\omega$ is a constant that depends on the wavelength and transmitter/receiver antennas gain \cite{a18}. 
Given the SNR $\gamma_{ij}$, the transmission rate of EU $i$ to edge node $j$ is defined as:
\begin{equation}
	 r_{i,j}^u = B_{ij} \log_2 (1+ \vartheta \gamma_{ij}) 
	\label{rate}
\end{equation}
where $\vartheta = -1.5/(\log (5 \cdot \text{BER})) $, as in \cite{a14}, and BER is the bit error rate target.   
By Substituting from (\ref{gamma}) in (\ref{rate}), the transmitted power can be written as
\begin{equation}
	 P^t_{ij} = \frac {N_0 \cdot B_{ij}}{g_{ij}} (2^{r_{ij}/B_{ij}}-1), 
	\label{pt}
\end{equation}
where $g_{ij}$ is the channel gain that is defined as 
\begin{equation}
g_{ij} = \vartheta \cdot \omega \cdot d_{ij}^{-\alpha} \cdot {|h_{ij}|}^2.  
\end{equation}
Accordingly, the total energy consumed by EU $i$ to send a data of length $\left| W_i \right|$ to edge node $j$ is
\begin{equation}
	 E_{ij} = \frac{P^t_{ij} \cdot \left| W_i \right|}{r_{ij}}=  \frac{\left| W_i \right| \cdot N_0 \cdot B_{ij} }{r_{ij} \cdot g_{ij}} (2^{r_{ij}/B_{ij}}-1).  
	\label{energy}
\end{equation} 
We remark that the equation in (\ref{energy}) considers the energy consumption  due to  the transmission, while ignoring the energy consumption due to processing since it is normally notable smaller, and it is neither affected by the EUs assignment nor the wireless resource allocation.  

\section{Joint EU Assignment and Resource Allocation \label{sec:sec4}}

Although leveraging hierarchical FL allows for avoiding data exchange between different EUs, the FL procedure can also drain all resources in the system. Thus, it is crucial to optimize the amount of resources used during the learning process, to avoid overloading the system, while maintaining the accuracy and convergence time on an acceptable level.  
This paper paves the way for optimizing the hierarchical FL performance in resource-constrained heterogeneous systems through answering the following questions: 
\begin{itemize}
	\item To which edge node should the participating EUs be allocated? 
	\item How can we optimally leverage the available amount of resources while adequately training the learning model?
\end{itemize} 
These questions are particularly important in our system, since we consider a real scenario of distributed nodes and IoT devices, where:   
(i) the global data distribution is imbalanced, and (ii) EUs' resources as well as edge computing resources are not as abundant as in the centralized server.   
Herein, we refer to ``resources'' as the the computation and communication resources, including time, energy, and wireless network bandwidth.  


\subsection{ Problem Formulation \label{sec:Formulation}}  

Typically, FL performance is measured by its convergence behavior, which depends on the divergence between the federated weights $\w_{\text{f}}^{b}$ and the central weights $\w_c^t$. Herein, the central weights represent the learning parameters that are calculated by the centralized gradient descent, when it considers the aggregated dataset of all EUs.  
This divergence represents the deviation caused due to the distribution of the datasets at different EUs while performing distributed learning.  Hence, it is always mandatory to obtain the minimal divergence in order to generate a closer model to the conventional centralized solution. 

Convergence bound of FedAvg or gradient-descent based FL on non-IID data has been studied in \cite{zhao_federated_2018, Convergence_FedAvg}. Such convergence analysis depicts two important facts:
\begin{itemize}
    \item there is always a trade-off between the communication-efficiency and convergence rate, and
    \item heterogeneity of the data slows down the convergence. 
\end{itemize}  
{
By extending the analysis in \cite{zhao_federated_2018} for hierarchical FL,  the upper bound of the divergence between federated weights and \textit{virtual} central weights can be approximated to 
\begin {equation}
\label{eq:divergence}
\|\w_{f} - \w_c \| \propto \sum_{j=1}^N  \sigma_{j} \times \|D^{(j)}\|_1, 
\end{equation} 
where $\sigma_{j}$ is the proportion of edge $j$ dataset to the global dataset, and $\|D^{(j)}\|_1$ is the classes distribution distance between the edge $j$ dataset and the global aggregated \textit{virtual} dataset. 
This upper bound can be obtained by adjusting the analysis in \cite{zhao_federated_2018} to follow the hierarchical FL settings while writing the upper bound as a function of $\|D^{(j)}\|$.  
Hence, in order to minimize the divergence between federated weights and \textit{virtual} central weights while accelerating FL convergence, it is crucial to maintain a balanced data distribution (similar to the global \textit{virtual} dataset) at the edge nodes  by  uniformly distributing the acquired data  across all edge nodes. This fact has been also proved mathematically in \cite{duan_self-balancing_2021}.  
} 

Based on the previous theoretical analysis, we formulate an optimization problem that considers the imbalanced data distribution and wireless communication resources to dynamically adapt  the EUs assignment to the edge nodes in real time, while minimizing the learning loss. Our problem aims at optimally allocating EUs to different edge nodes, such that it obtains a balanced data distribution at different edge nodes, hence accelerating the FL convergence.   
To achieve this, we opt to minimize the Kullback–Leibler divergence (KLD) of the data distributions \cite{kullback} in order to tackle the global imbalanced data problem.  
KLD, also called relative entropy, measures how a probability distribution $H(c_k)$ is different from a second, reference probability distribution $Q(c_k)$ over the same random variable $c_k$.    
Let $H_j(c_k)$ and $Q(c_k)$ be the data classes distribution at edge node $j$ and uniform distribution of a discrete random variable $\mathcal{C}$ with possible values $\mathcal{C} = \left\{c_1, c_2, \cdots, c_K \right\}$, respectively. 
The KLD is defined as 
\begin {equation}
\label{eq:KLD}
D_{KL} (H_j \left|\right| Q) = \sum_{k=1}^K  H_j(c_k) \log {\frac{H_j(c_k)}{Q(c_k)}},
\end{equation} 
where $\sum_{k=1}^K H_j(c_k) = 1$ and $\sum_{k=1}^K Q(c_k)=1$, as well as $H_j(c_k) > 0$ and $Q(c_k) > 0$, for any $c_k \in \mathcal{C}$, such that $\mathcal{C}$  is the set of all possible classes available at all EUs.     
Herein, $H_j(c_k)$ represents the actual data distribution at the edge, while $Q(c_k)$ representing the ideal data distribution at different edge nodes (i.e., uniform distribution or the global \textit{virtual} dataset distribution).  
A KLD of zero indicates that the data classes are uniformly distributed on all edge nodes, i.e., the acquired data at different edge nodes is IID.    
%

Thus, the main objective of our optimization problem is to optimally allocate the $M$ EUs to the available $N$ edge nodes, such that the KLD is minimized. However, given the wireless communication constraints, distributing the EUs to different edge nodes will not be an easy task, especially when considering the non-IID data distribution between different EUs. Thus, our optimization problem is formulated as follows:

\begin{eqnarray}
\mbox{\bf P1:} &	&  \min_{\lambda_{ij}, B_{ij}} \  \sum_{j=1}^{N}  D_{KL} (H_j \left|\right| Q)     
	\label{eq:optimize_prob} \\
\mbox{s.t.} \nonumber\\ 
&&	 T_{i}^c  + \sum_{j=1}^{N} \lambda_{ij} \cdot  {L_{ij}}  \le T^{m},  \ \    \forall i \in  \mathcal{M}   \label{eq:C1} \\ 
&&  \sum_{j=1}^{N} \lambda_{ij} \cdot  E_{ij}  \le E_i^m, \ \       \forall i \in \mathcal{M}         \label{eq:C2} \\ 
&&  \sum_{i=1}^{M} \lambda_{ij} \cdot  {B_{ij}}  \le B_j^{m},   \ \     \forall j \in \mathcal{N}   \label{eq:C3} 		
\end{eqnarray} 
\begin{eqnarray}
	&& \sum_{j=1}^{N} \lambda_{ij} = 1, \ \       \forall i \in  \mathcal{M}         \label{eq:C4}  \\
	&&  \lambda_{ij} = \left\{0, 1\right\},  \ \ \  \forall i \in \mathcal{M}  \  \& \  \forall j \in \mathcal{N}   \label{eq:C5}
\end{eqnarray} 
where $\lambda_{ij}$ is the edge indicator, such that $\lambda_{ij}=1$ when EU $i$ is allocated to edge node $j$.    
The constraint in (\ref{eq:C1}) ensures that the transmitted updates from all EUs will be received at the edge nodes with a maximum delay $T^{m}$. This constraint considers the computational delay, at each EU $T_{i}^c$, in addition to the transmission delay $L_{ij}$ from an EU $i$ to an edge $j$, where  $L_{ij} = \frac{\left| W_i \right|}{r_{i,j}^u} +	\xi_i$. 
The constraint in (\ref{eq:C2}) considers the energy budget/limit at different battery-operated EUs. Hence, it ensures that the energy consumed by EU $i$ to send its local updates to edge $j$ cannot exceed the maximum transmission energy $E_i^m$.   
The constraint in (\ref{eq:C3}) refers to the network capacity constraint, where $B_{ij}$ is the maximum fraction of bandwidth $B_j^{m}$ that can be used by EU $i$ to communicate with edge $j$. We remark that $B_{ij}$ depends on the number of EUs communicating with the edge, and it is notified by the edge nodes to the EUs.  
The constraints in (\ref{eq:C4}) and (\ref{eq:C5}) ensure that each EU will be connected with only one edge.   

The optimization variables in (\ref{eq:optimize_prob}) are the $\lambda_{ij}$'s and $B_{ij}$'s, i.e., each EU needs to be assigned to an edge node, and each edge node needs to allocate the bandwidth for the assigned  EUs. Meanwhile, the proposed optimization problem assigns the EUs to the available edge nodes such that the edge nodes' data distributions turn to be the closest to the uniform distribution. 
In other words, our optimization problem aims to adjust the EU allocation at different edge nodes in order to obtain balanced data distribution over all edge nodes (by optimizing the KLD). This balanced data (or class) distribution, from a practical perspective, will lead to accelerating the learning process and promoting the swift convergence (as will be shown in our results).    
 
Looking at problem formulation in (\ref{eq:optimize_prob}), one can see that it is an integer programming problem, which is NP-complete problem \cite{a15}.  
Also, the well-known approaches of converting the problem into a convex problem or a Geometric Program (GP), would not work in this case due to the existence of the constraints in (\ref{eq:C4}) and (\ref{eq:C5}) with the non-linearity of the objective function (according to the definition of KLD in (\ref{eq:KLD})).   
Thus, below we envision a methodology to solve this problem.


\subsection{Proposed Solution  \label{sec:Solution}}

To solve the formulated problem in (\ref{eq:optimize_prob}), we first reformulate the NP-complete optimization problem to have a linear objective function; then we \textit{relax} it to a Linear Program (LP) by removing the bandwidth constraint, and solve the linear program.  
After that, we \textit{round} the LP solution to a solution that satisfies the original integer constraint. Finally, given the LP solution, we optimize the bandwidth allocation to the available EUs, such that we maintain the KLD minimization obtained by the LP solution.  
In what follows, we illustrate the proposed solution in details.
      
First, we rewrite the objective function as follows:
\begin{eqnarray}
\label{eq:obj1} 
Z &=& \min_{\lambda_{ij}, B_{ij}} \  \sum_{j=1}^{N} D_{KL} (H_j \left|\right| Q)  \nonumber\\ 
  &=& \min_{\lambda_{ij}, B_{ij}} \  \sum_{j=1}^{N} \sum_{k=1}^K  H_j \log {\frac{H_j}{Q}},   \nonumber\\
  &=& \min_{\lambda_{ij}, B_{ij}} \left( \sum_{j=1}^{N} \sum_{k=1}^K  H_j \log {\left(H_j\right)} - \sum_{j=1}^{N}  \sum_{k=1}^K  H_j \log {\left(Q\right)} \right).    \nonumber\\ 
\end{eqnarray}	
For the sake of brevity, we refer to $H_j(c_k)$ by $H_j$, and $Q(c_k)$ by $Q$. Given that $Q$ is representing the uniform distribution, i.e., $Q$ is fixed, Hence   
\begin{eqnarray}
\label{eq:obj2} 
Z &=& \min_{\lambda_{ij}, B_{ij}} \left( \sum_{j=1}^{N} \sum_{k=1}^K  H_j \log {\left(H_j\right)} - \log {\left(Q\right)} \sum_{j=1}^{N}  \sum_{k=1}^K  H_j \right),   \nonumber\\ 
  &=& \min_{\lambda_{ij}, B_{ij}} \left( \sum_{j=1}^{N} \sum_{k=1}^K  H_j \log {\left(H_j\right)} - N \cdot \log {\left(Q\right)} \right).   
\end{eqnarray} 
Given that the term $N \cdot \log {\left(Q\right)}$ is fixed with respect to $\lambda_{ij}$ and $B_{ij}$, thus the objective function can be written as 
\begin{eqnarray}
\label{eq:obj3} 
Z  &=& \min_{\lambda_{ij}, B_{ij}} \left( \sum_{j=1}^{N} \sum_{k=1}^K  H_j \log {\left(H_j\right)}  \right)
    = \max_{\lambda_{ij}, B_{ij}} \left( \sum_{j=1}^{N} \chi_j (\mathcal{C})  \right).    \nonumber\\
\end{eqnarray}
Now, the objective function in (\ref{eq:obj3}) is turned to be equivalent to maximizing the information entropy (or Shannon entropy) $\chi_j$ at the edge nodes \cite{Information_Theory}, where $\chi_j (\mathcal{C}) = - \sum_{k=1}^K  H_j(c_k) \log {H_j(c_k)} $. It can be easily proved that the maximum entropy can be only achieved at the value associated with uniform distribution, which can be maintained by distributing all data classes equally on all edge nodes.      
This implies that $H_j$'s should have the same value for all $j \in \mathcal{N}$.      
However, $H_j$ depends on: (i) $\lambda_{ij}$'s, i.e., the allocated EUs at edge node $j$, and (ii) $c_{k}^i$'s, i.e., the associated data classes for the EUs.     
Hence, $H_j(c_k)$ can be defined as
  \begin {equation}
\label{eq:H}
H_j(c_k) =   {\frac{\sum_{i=1}^M \lambda_{ij} \cdot c_k^i}{\sum_{k=1}^K \sum_{i=1}^M \lambda_{ij} \cdot c_k^i}}. 
\end{equation}   
Given that the maximum entropy is achieved with uniform distribution, our  objective function can be written as:      
\begin {equation}
\label{eq:obj4}
Z  = \min_{\lambda_{ij}, B_{ij}} \left(\sum_{k=1}^K  \sum_{ \left\{j, \tilde{j} \right\} \in \mathcal{S}} \left|\sum_{i=1}^M \lambda_{ij} \cdot c_k^i - \sum_{i=1}^M \lambda_{i \tilde{j}} \cdot c_k^i\right| \right), 
\end{equation}
where $\mathcal{S}$ is a subset of all possible subsets that can be selected from $\mathcal{M}$ without replacement. For example, if we have 3 edge nodes, the set $\mathcal{S}$ can be defined as $\mathcal{S} = \left\{ \left\{ 1,2 \right\}, \left\{ 1,3 \right\}, \left\{ 2,3 \right\} \right\}$.

After simplifying our objective function, we still have the non-linear constraint in (\ref{eq:C1}), due to the dependence of the transmission rate on the bandwidth, as well as the integer constraint in (\ref{eq:C5}). Thus, to convert the formulated optimization problem into a LP problem  that can be easily solved using many well-known techniques   \cite{linear_programming}, we reformulate the problem as follows: 
\begin{eqnarray}
\mbox{\bf P2:} &	&  \min_{\lambda_{ij}} \left(\sum_{k=1}^K  \sum_{ \left\{j, \tilde{j} \right\} \in \mathcal{S}} \left|\sum_{i=1}^M \lambda_{ij} \cdot c_k^i - \sum_{i=1}^M \lambda_{i \tilde{j}} \cdot c_k^i\right| \right) 
	\label{eq:optimize_prob2} \\
\mbox{s.t.} \nonumber\\
&&	 T_{i}^c  + \sum_{j=1}^{N} \lambda_{ij} \cdot  {L_{ij}}  \le T^{m},  \ \    \forall i \in  \mathcal{M}   \label{eq:C21} \\ 
&&  \sum_{j=1}^{N} \lambda_{ij} \cdot  E_{ij}  \le E_i^m, \ \       \forall i \in \mathcal{M}         \label{eq:C22} \\  
	&& \sum_{j=1}^{N} \lambda_{ij} = 1, \ \       \forall i \in  \mathcal{M}         \label{eq:C24}  \\
	&& 0 \leq \lambda_{ij} \leq 1, \ \forall i \in \mathcal{M}  \  \& \  \forall j \in \mathcal{N}   \label{eq:C25}
\end{eqnarray}

Now, to solve the original formulated problem in (\ref{eq:optimize_prob}) and obtain the best assignment to the edge nodes with bandwidth allocation for all EUs, we propose an efficient EUs Assignment and Resource Allocation (EARA) algorithm (see Algorithm \ref{alg:alg1}).      
%
The proposed EARA algorithm aims to solve our problem in two steps.  
Initially, it is assumed that each edge will have enough bandwidth to be allocated equally to all associated EUs.  
Then, in the first step, the problem in (\ref{eq:optimize_prob2}) is solved to obtain $\lambda_{ij}$. However, in (\ref{eq:optimize_prob2}), the integer constraint of $\lambda_{ij}$ is relaxed, while assuming equal bandwidth allocation for all EUs. Thus, the obtained $\lambda_{ij}$ is rounded to ''0'' or ''1'', such that the original constraints in (\ref{eq:C4}) and (\ref{eq:C5}) are maintained. 
We remark here that the proposed EARA algorithm allows for enhancing the hierarchical FL performance by enabling the EUs to have two configuration options: Single Connectivity Allocation (SCA), and Dual Connectivity Allocation (DCA). 
The former refers to assigning each EU to only one edge, while the later refers to the possibility of an EU to be connected to one or two edge nodes simultaneously, leveraging Dual Connectivity (DC)\footnote{In Release 12 of the 3GPP LTE specifications, DC is initially designed to enable the users to connect with two cells at the same time, preferably in heterogeneous networks \cite{Dual_Conn_standart}. Then, the required non-standalone deployment of 5G technology has extended the concept of DC to be used with multiple radio technologies, and in particular, for connecting 5G cells to a 4G core network.} and multicast transmission \cite{Dual_Conn_standart, Multicast}.   
We remark that DC is one of the substantial technologies adopted by 5G to increase the network throughput and mobility robustness by enabling the users to simultaneously connect with macro and small-cell base stations \cite{PA_DC}.  In addition, multicast transmission
provides significant bandwidth savings for sending duplicate model update from one EU to two edge nodes on a same resource share.  
It is important also to note that DCA configuration represents an extension to our problem formulation, since it relax the constraint in (\ref{eq:C24}) to allow the EU to connect to one or two edge nodes simultaneously.  

Using SCA configuration, we set $\lambda_{ij}^*=1$, where
\begin {equation}
\label{eq:lamdaMAx}
\lambda_{ij}^* =  \arg\max_{j \in \mathcal{N}} \left ( \lambda_{ij} \right), 
\end{equation}
while setting the others $\lambda_{ij}, \forall j \in \mathcal{N} \ \backslash \left\{ \lambda_{ij}^* \right\}$ to $0$. 
In DCA configuration, we consider the maximum two values of $\lambda_{ij}$'s, i.e., $\lambda_{ij}^1=\arg\max_{j \in \mathcal{N}} \left ( \lambda_{ij} \right)$,  and $\lambda_{ij}^2= \arg\max_{j \in \mathcal{N} \ \backslash \left\{ \lambda_{ij}^1 \right\}} \left ( \lambda_{ij} \right)$, such that: 
(i) $\lambda_{ij}^1$ is set to 1, and 
(ii) $\lambda_{ij}^2$ is set to 1, only if $\lambda_{ij}^2 > \nu$, while setting the others $\lambda_{ij}, \forall j \in \mathcal{N} \ \backslash \left\{ \lambda_{ij}^1, \lambda_{ij}^2 \right\}$ to $0$.  
Herein, $\nu$ is a predefined threshold that represents the EU ability to connect to two edge nodes.   

In the second step, the available bandwidth $B_j^{m}$ of an edge node $j$ is allocated to the associated EUs based on their importance.           
Herein, \textit{importance} of an EU refers to its weight (or role) in minimizing the KLD. For instance, EUs with data classes that are different from the available ones at edge node $j$ will be weighted more than others, in order to maintain the KLD minimization.  
In this context, an edge $j$ ranks all EUs $M_j$ that are assigned to it, after solving (\ref{eq:optimize_prob2}). Then, it allocates the minimum bandwidth required for the most important  EU to satisfy its latency constraint in (\ref{eq:C1}). This bandwidth allocation process is repeated until all EUs receive their bandwidth allocation or the edge node consumes the available bandwidth budget, i.e., the original constraint in (\ref{eq:C3}) is maintained.            

\begin{algorithm}
\caption{EUs Assignment and Resource Allocation (EARA) Algorithm}
\label{alg:alg1}
\begin{algorithmic}[1]
\State {\textbf{Input:} $B_{ij} = B_{f}$, where $B_{f}$ is the fixed bandwidth assigned to EU $i$ at edge $j$. } \\ 
{\underline{\textbf{At the centralized server:} (\textit{EUs assignment})} } 

\State {Solve the problem in (\ref{eq:optimize_prob2}) to obtain $\lambda_{ij}$.  } 
\If {SCA is ON}
\State {Set $\lambda_{ij}^* = 1$, while setting $\lambda_{ij} = 0, \ \forall j \in \mathcal{N} \ \backslash \left\{ \lambda_{ij}^* \right\} $.  } 
\EndIf
\If {DCA is ON}
\State {Set $\lambda_{ij}^1=1$.}
\If {$\lambda_{ij}^2 > \nu$} \State {$\lambda_{ij}^2 = 1.$} \Else  \State{$\lambda_{ij}^2 = 0$.} 
\EndIf
\State {Set $\lambda_{ij} = 0, \ \  \forall j \in \mathcal{N} \ \backslash \left\{ \lambda_{ij}^1, \lambda_{ij}^2 \right\} $.} 
\EndIf
\State {Send the values of $\lambda_{ij}$'s to all edge nodes. } \\

---------------------------------------------------------------------\\

{\underline{\textbf{At the Edge:} (\textit{bandwidth allocation})} } 

\State {Receive $\lambda_{ij}$'s from the centralized server .}
\State {Sort the assigned EUs based on their weights in an descending order. }
\For {$i = 1 \to M_j$}
\State {Calculate the minimum $B_{ij}$ that satisfies the constraint in (\ref{eq:C1}). }
\If {  $\sum_{i=1}^{M_j} \lambda_{ij} \cdot  {B_{ij}} \geq B_j^{m} $ }
\State{Break}  \Comment{the available bandwidth is consumed.} 
\EndIf
\EndFor
\State \Return{ $\lambda_{ij}$, $B_{ij}$. }
\end{algorithmic}
\end{algorithm}



\section{Performance Evaluation \label{sec:sec5}} 

In this section, we first present the simulation environment that is used to derive our results. Then, we assess the performance of the proposed  hierarchical FL framework compared to  state-of-the-art techniques.  In particular, our results assess the performance of the proposed EARA algorithm using two real-world healthcare datasets to measure the enhancement in the learning speed and accuracy. 

\subsection{Environment setup \label{sec:setup}}

For our performance evaluation, we use two datasets, namely,  \textit{Heartbeat} dataset and \textit{Seizure} dataset. 
\textit{Heartbeat} dataset in \cite{Heartbeat_dataset} includes groups of heartbeat signals, i.e., electrocardiogram (ECG) signals, derived from MIT-BIH Arrhythmia dataset \cite{MIT_dataset} for heartbeat classification. These signals represent the normal case and the cases affected by different arrhythmias and myocardial infarction, hence they are used in exploring heartbeat classification using deep neural network architecture.      
\textit{Seizure} dataset represents electroencephalogram (EEG) measurements comprised of a mixture of waveforms recorded over time that reflect the electrical activity of the brain and  obtained from electrodes placed on the scalp, typically in patients with epilepsy. The seizure dataset consists of EEG recordings in such patients that was acquired on a Nicolet 
machine at the American University of Beirut Medical Center (AUBMC) using the international 10-20 system for electrode placement. The occurrence, number and timing of seizure(s) in each EEG recording was annotated by an experienced electroencephalographer using established criteria that can be summarized as an abrupt change in frequency or amplitude of 
the waveforms that exhibit evolution in time or space and lasting for 10 seconds or more \cite{AUB_dataset}.  

For the distributed settings, we consider different number of EUs while the data is distributed randomly into the EUs, such that we maintain non-IID data distribution between different EUs.  
{
For the \textit{Seizure} dataset, the number of possible classes is $3$. Hence, we consider $3$ edge nodes with $13$ users. These users are initially assigned to the edge nodes such that the edge nodes level distribution are obtained as in Table \ref{tab:s}. 
\begin{table}[]
\renewcommand{\arraystretch}{1.2}
\caption{Initial edge-level distribution for the \textit{Seizure} dataset.}
\label{tab:s}
\centering
\begin{tabular}{llll}
                          & \multicolumn{3}{l}{Instances per Class}                      \\ \hline
\multicolumn{1}{l|}{Edge} & \multicolumn{1}{l|}{class 0}    & \multicolumn{1}{l|}{class 1}    & class 2    \\ \cline{2-4} 
\multicolumn{1}{l|}{0}    & \multicolumn{1}{l|}{1459} & \multicolumn{1}{l|}{25}   & 25   \\\hline
\multicolumn{1}{l|}{1}    & \multicolumn{1}{l|}{25}   & \multicolumn{1}{l|}{1160} & 25   \\\hline
\multicolumn{1}{l|}{2}    & \multicolumn{1}{l|}{25}   & \multicolumn{1}{l|}{25}   & 1238  \\ \hline
\end{tabular}
\end{table}
As for the  \textit{Heartbeat} dataset, the number of classes is $5$. Hence, we consider $5$ edge nodes with $18$ users. These users are initially assigned to the edge nodes such that each edge contains the distribution illustrated in Table \ref{tab:hb}. 
\begin{table}[]
\centering
\renewcommand{\arraystretch}{1.2}
\caption{Initial edge-level distribution for the  \textit{Heartbeat} dataset.}
\label{tab:hb}
\begin{tabular}{llllll}
                       & \multicolumn{5}{l}{Instances per Class$\times10^{3}$}                                                                    \\ \hline
\multicolumn{1}{l|}{Edge} & \multicolumn{1}{l|}{class 0} & \multicolumn{1}{l|}{class 1} & \multicolumn{1}{l|}{class 2} & \multicolumn{1}{l|}{class 3} & class 4 \\ \cline{2-6} 
\multicolumn{1}{l|}{0} & \multicolumn{1}{l|}{10} & \multicolumn{1}{l|}{10} & \multicolumn{1}{l|}{0}  & \multicolumn{1}{l|}{0}  & 0  \\\hline
\multicolumn{1}{l|}{1} & \multicolumn{1}{l|}{0}  & \multicolumn{1}{l|}{0}  & \multicolumn{1}{l|}{10} & \multicolumn{1}{l|}{10} & 0  \\\hline
\multicolumn{1}{l|}{2} & \multicolumn{1}{l|}{10} & \multicolumn{1}{l|}{0}  & \multicolumn{1}{l|}{0}  & \multicolumn{1}{l|}{0}  & 10 \\\hline
\multicolumn{1}{l|}{3} & \multicolumn{1}{l|}{0}  & \multicolumn{1}{l|}{10} & \multicolumn{1}{l|}{10} & \multicolumn{1}{l|}{0}  & 0  \\\hline
\multicolumn{1}{l|}{4} & \multicolumn{1}{l|}{0}  & \multicolumn{1}{l|}{0}  & \multicolumn{1}{l|}{0}  & \multicolumn{1}{l|}{10} & 10  \\ \hline
\end{tabular}
\end{table}

\textbf{Models architecture:} 
For the  \textit{Heartbeat} dataset, we use the model presented in \cite{EMina},  which expects $1$ input channel and outputs probabilities for $5$ classes. For the \textit{Seizure} dataset, we also use a similar model, but we adapted it to accommodate the $19$ input channels and the $3$ output classes. Our models' details can be found in \cite{models_repo}. 

The training-related parameters are the same for both datasets, and they are: local epochs equal $1$, local batch size equals $10$, and learning rate equals $0.001$, while selecting ADAM optimizer. 
For the centralized baseline, it is assumed that the batch size is $50$ for the \textit{Heartbeat} dataset and $30$ for the \textit{Seizure} dataset.  This is because the local batch size should be multiplied by the number of edge nodes, so that at each communication round in the federated settings, the same number of data points are used for both the central and federated models \cite{zhao_federated_2018}. 
}


\subsection{Simulation results \label{sec:sim_results}}

In what follows, we compare the proposed EARA algorithm with its two configuration options, i.e., SCA and DCA, against the state-of-the-art solution, such as  \cite{Client-Edge-Cloud, HFL_sparse}, hereinafter we referred to it as Distance-based Allocation (DBA). The DBA scheme considers the same hierarchical FL architecture in Figure~\ref{fig:system_model}, while assigning different EUs to the nearest edge node, based on the distance between the EUs and the edge nodes. Furthermore,  we compare our EARA algorithm against a baseline approach, namely, the centralized learning, which represents the benchmark of the FL performance since it assumes that all EUs' data are collected at a centralized server for the training.         

\begin{figure}[t!]
	\centering
		\scalebox{1.55}{\includegraphics[width=0.27 \textwidth]{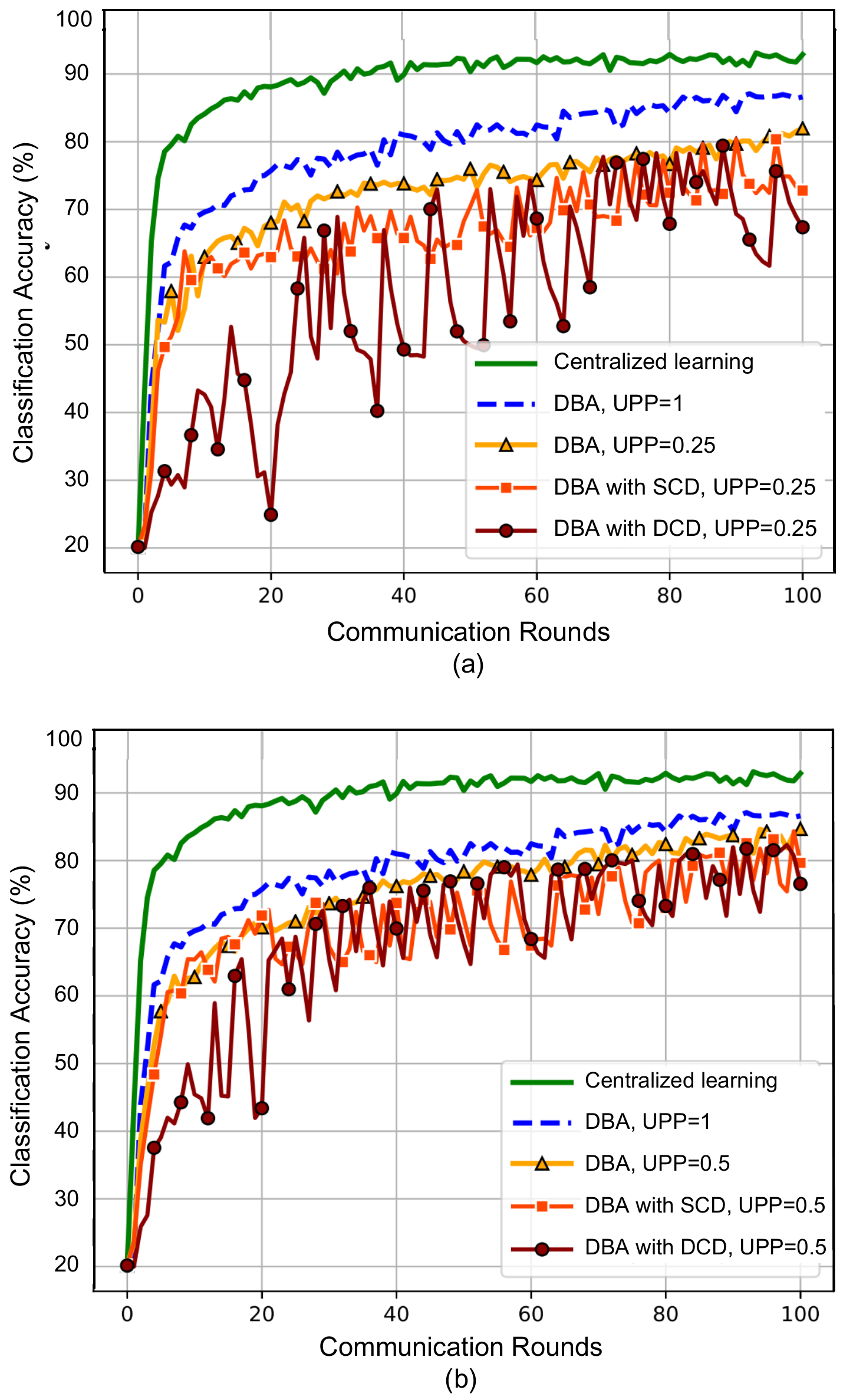}}
	\caption{Effect of varying UPP on the obtained classification accuracy for DBA scheme, using \textit{Heartbeat} dataset.  }
	\label{fig:users_selection}
\end{figure}

The first aspect we investigate is the influence of the EUs dropping on the obtained classification accuracy. To this end, Figure~\ref{fig:users_selection}  presents the classification accuracy as a function of the communication rounds between the edge nodes and the centralized server, while considering the DBA scheme.   
In this figure, we consider different values for Users Participating Percentage (UPP), which represents the percentage of the EUs that the edge nodes receive their updates during the FL training process. Also, we consider two extreme cases, i.e., Single Class Dropping (SCD) and Dual Classes Dropping (DCD). The former refers to missing the whole samples of one data class by dropping their EUs, while the latter refers to missing the samples of two data classes. 
It is possible to observe that DBA scheme has significantly less performance than the centralized learning, even when considering all EUs' updates, i.e., UPP=1.  An intuitive explanation is that the non-IID data on different edge nodes increases the bias in the learned pattern by the centralized FL server, which results in low prediction accuracy. This confirms the empirical observations in \cite{Convergence_FedAvg} that heterogeneity of the data harms the accuracy and slows down the convergence.     
More interestingly, decreasing the number of participating EUs in the training yields substantial reduction in the obtained accuracy, especially when the dropping of the EUs leads to missing the samples of one or two data classes.  Thus, it is crucial to address the heterogeneity of the data at different edge nodes, to reduce the effect of non-IID data, while considering all EUs with important/unique data.   
These results are the main motivation behind the proposed EARA scheme. 

\begin{figure}[t!]
	\centering
		\scalebox{1.5}{\includegraphics[width=0.27 \textwidth]{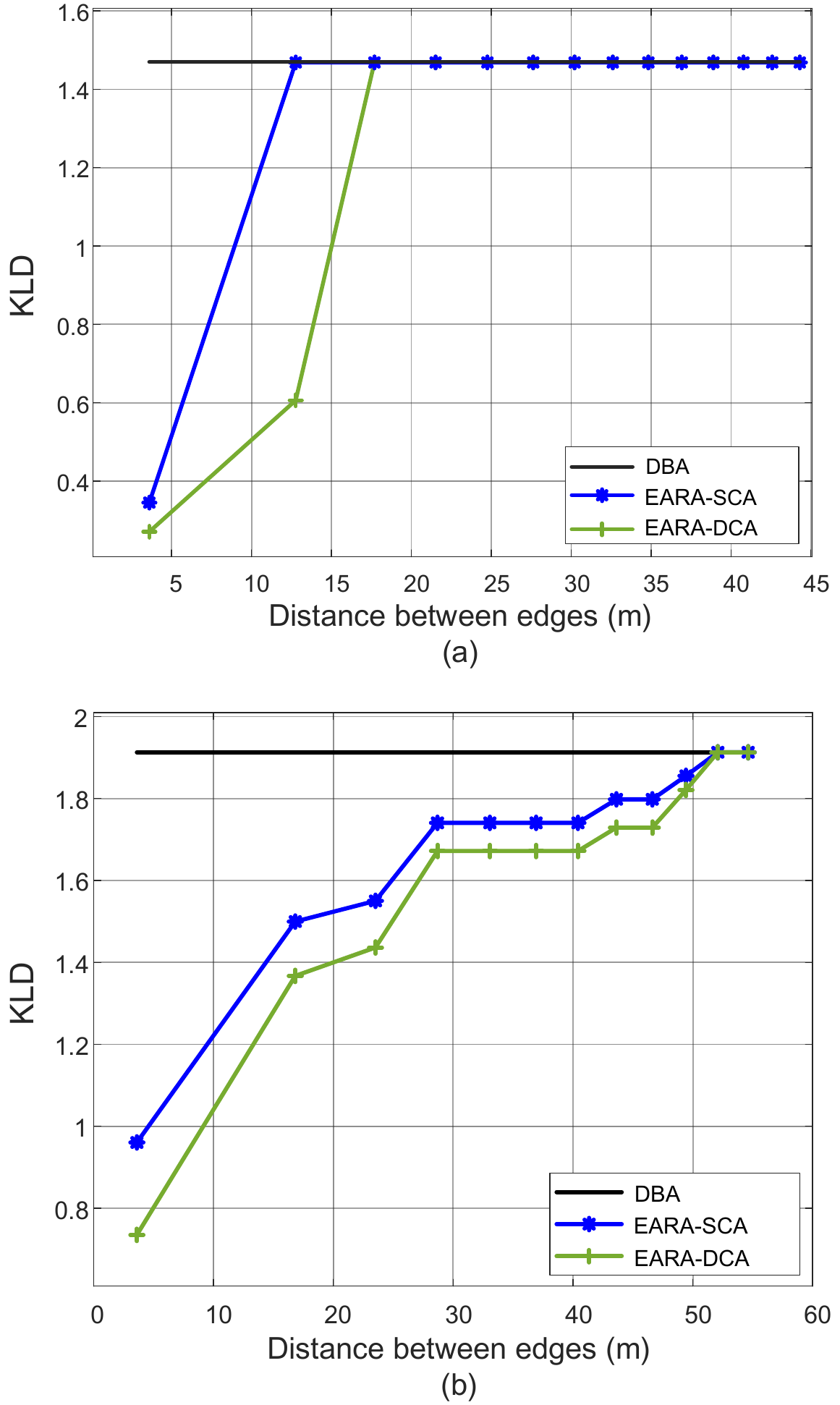}}
	\caption{KLD variations with increasing the distance between EUs and edge nodes, for different EUs assignment strategies using \textit{Heartbeat} dataset, while considering: (a) 3 edge nodes and 13 EUs, (b) 5 edge nodes and 18 EUs.  }
	\label{fig:KLD compare}
\end{figure}

The second aspect we are interested in is how the proposed EARA algorithm influences the performance of the hierarchical FL.  Figure~\ref{fig:KLD compare} compares the proposed EARA algorithm with its two configuration options against the DBA approach.  
This figure depicts the KLD variations of all edge nodes (i.e.,  $\sum_{j=1}^{N}  D_{KL} (H_j \left|\right| Q) $) while varying the distance between EUs and edge nodes.      
It is clear that our EARA algorithm consistently outperforms the DBA approach. This figure indicates that the proposed EARA algorithm decreases the local data imbalance unlike DBA that obtains the highest KLD, while the dual connectivity option in EARA algorithm (EARA-DCA) obtaining the lowest KLD.  However, with increasing the distance between EUs and edge nodes, the energy consumption increases significantly for the EUs that are assigned to an edge node other than the nearest one. This violates the energy constraint in (\ref{eq:C2}), at large distances, hence the EUs are assigned to the nearest edge nodes, which results in converging the EARA performance to the performance of the DBA.     
In short, Figure~\ref{fig:KLD compare} suggests that the edge nodes can achieve better balanced data when more EUs participate in training or when the EUs are assigned uniformly to the edge nodes. 

\begin{figure}[t!]
	\centering
		\scalebox{1.67}{\includegraphics[width=0.27 \textwidth]{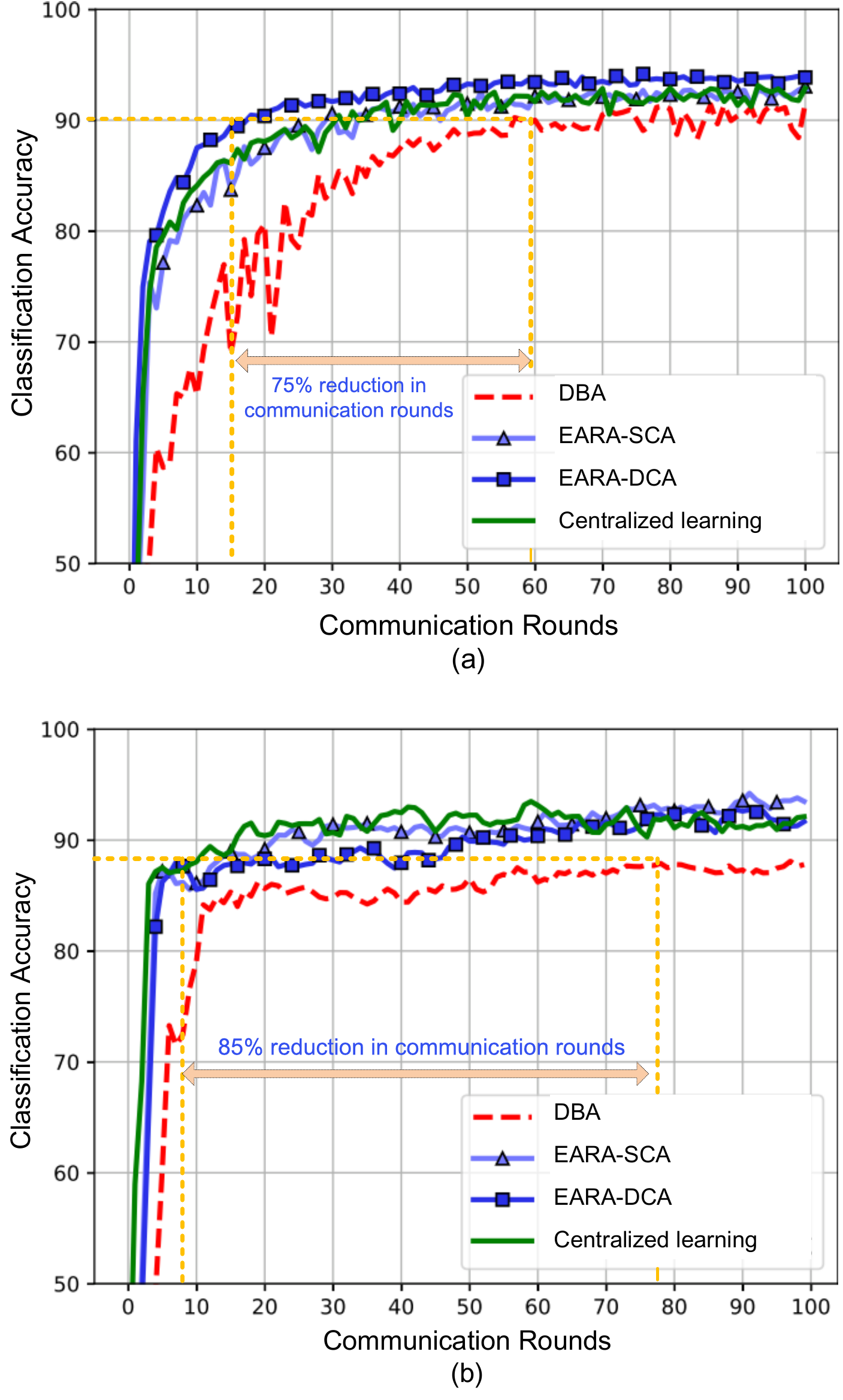}}
	\caption{Classification accuracy as a function of the communication rounds, for different EUs assignment strategies: (a) using \textit{Heartbeat} dataset, (b) using \textit{Seizure} dataset.  }
	\label{fig:convergence}
\end{figure}

The better performance of the proposed EARA algorithm in decreasing the local data imbalance leads to a significant enhancement in the obtained classification accuracy, as shown in Figure~\ref{fig:convergence}. Each marker therein corresponds to an EUs assignment strategy, and its x- and y- coordinates, respectively, correspond to the communication rounds and the achieved classification accuracy. 
This figure highlights how the performance of our algorithm approaches  the centralized benchmark solution, while reducing the communication rounds between the edge nodes and the centralized server by $75-85\%$ compared to DBA approach, for the same model accuracy.   
This is because of the adopted EUs assignment strategy that allows for maintaining balanced data distributions at different edge nodes. 
We highlight here that obtaining balanced data distribution or less-skewed data is an important issue to avoid the weights divergence of the learning model. 
Instead, the DBA approach obtains a slower convergence, while saturating at a lower accuracy value compared to our EARA algorithm due to the  weights divergence at the edge nodes that are affected by their local skewed data distributions.  
More interestingly, leveraging EARA algorithm with dual connectivity option (EARA-DCA) allows for sending duplicate updates to two edge nodes concurrently, hence reducing more the effect of imbalanced data, which leads to an out-performance even compared to the centralized learning benchmark.  
This resembles the well-known machine learning fact that the accuracy improves as the scale of the training expands.   

Figure~\ref{fig:convergence} highlights also that the differences in the obtained accuracy, leveraging different EUs assignment schemes, are obvious when considering \textit{Heartbeat} dataset. This is because \textit{Heartbeat} dataset has a more challenging classification problem compared to \textit{Seizure} dataset; hence, the imbalanced data has a major effect on the obtained performance.    
Moreover, we remark that our focus in this paper is to minimize the gap between the hierarchical FL performance and the centralized learning performance, instead of obtaining the most accurate learning model. Hence, the adopted deep learning models, described earlier, are sufficient for this purpose. 

\begin{figure}[t!]
	\centering
		\scalebox{1.5}{\includegraphics[width=0.27 \textwidth]{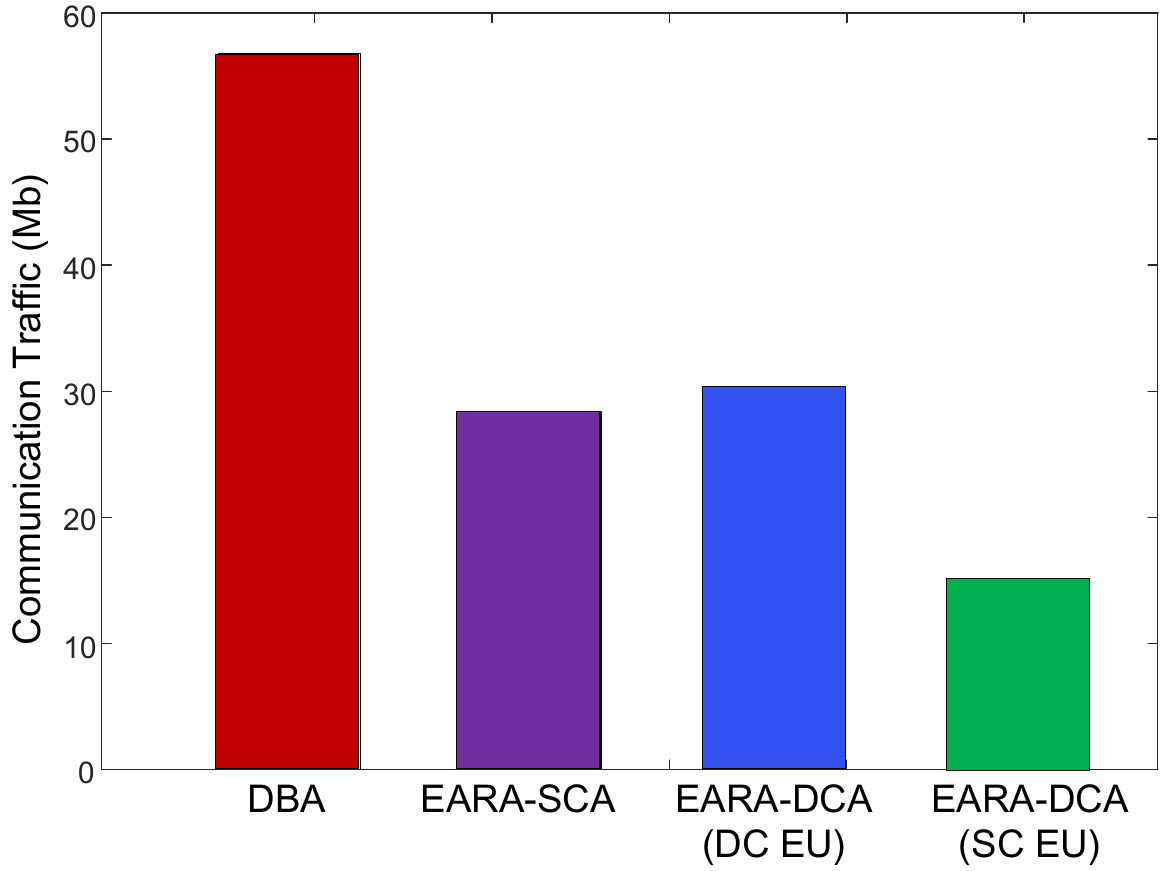}}
	\caption{Communication traffic per EU for different EUs assignment strategies, using \textit{Heartbeat} dataset.  }
	\label{fig:Traffic}
\end{figure}

Finally, in Figure~\ref{fig:Traffic}, we assess the communication traffic load per EU for our EARA approach compared to the DBA approach, while achieving $90\%$ classification accuracy on \textit{Heartbeat} dataset.  This figure depicts that the proposed approach is communication-efficient since our EARA-SCA reduces the communication traffic by $50\%$ compared to DBA. For EARA-DCA, it significantly reduces the communication rounds, hence  it reduces the communication traffic by $73\%$ for EUs that have been assigned to one edge, i.e., single-connectivity EUs (SC EU), compared to DBA. For EUs that have been assigned to two edge nodes, i.e., dual-connectivity EUs (DC EU), the communication traffic is increased by $3\%$ compared to EARA-SCA, while still obtaining less communication traffic by $47\%$ compared to DBA.       
Herein, the communication traffic is calculated considering the deep learning models, described earlier, with 14,789 parameters, while assuming that each parameter is represented by 4 bytes, as in the PyTorch tutorial \cite{Pytorch_training}. Moreover, we remark that by leveraging the recent technologies adopted by 5G, two edge nodes (such as macro cell base station and small cell base station) can simultaneously transmit the data to one EU with the help of millimeter-wave (mmWave) massive MIMO technology \cite{ML_DC}. This can significantly decrease the overhead resulting from the dual connectivity.    

\section{Conclusion\label{sec:conclusion}}

In this paper, we propose an efficient user assignment and resource allocation scheme for hierarchical FL. The proposed scheme allows for leveraging the massive data generated from IoT devices for training deep learning models, while effectively addressing the challenges and requirements posed by the data privacy and resource-constrained environment. 
In particular, the proposed scheme  allows for rendering the distribution of EUs' data, at the edge, to be close to the uniform distribution, which significantly reduces the communication rounds between the edge nodes and the centralized server.  
The proposed scheme could significantly alleviate the impact of local imbalanced and non-IID data, and reduce the KLD compared to the state-of-the-art solutions that rely on distance-based user assignment. 
Hence, it accelerates the learning process and reduces the communication traffic by decreasing the global aggregation rounds on the centralized server.  
Our results show that the proposed approach can reduce $75-85\%$ of the  communication rounds between the edge nodes and the centralized server, compared to the state-of-the-art solutions, while preserving the same model accuracy.  

The future research directions can include developing online algorithms to select a subset of outstanding users that accelerate the hierarchical FL process, while minimizing the communication overheads. Moreover, the tight synchronization needed among nodes in FL and the presence of a centralized server are  still challenging problems that require further research to make FL techniques suitable for highly dynamic scenarios like vehicular applications.  

\section*{Acknowledgement}
This work was made possible by NPRP grant \# NPRP12S-0305-190231 from the Qatar National Research Fund (a member of Qatar Foundation). The findings achieved herein are solely the responsibility of the authors.

\balance 

\bibliographystyle{IEEEtrannames}
\bibliography{FL_For_HC}

\end{document}